\documentclass[11pt,a4paper]{article}
\usepackage[hyperref]{acl2020}
\usepackage{times}
\usepackage{latexsym}
\usepackage{graphicx}
\usepackage{multirow}
\usepackage{comment}
\usepackage{subcaption}
\usepackage{booktabs}

\usepackage{microtype}

\aclfinalcopy 
\def\aclpaperid{9} 

\usepackage{xcolor}
\usepackage{xspace}
\usepackage{xargs}

\title{Language Models as Fact Checkers?}

\author{
Nayeon Lee$^1$\thanks{\hspace{2pt} Work done while at Facebook AI.}\quad\quad Belinda Z. Li$^2$\quad\quad Sinong Wang$^2$ \\
\bf Wen-tau Yih$^2$\quad\quad Hao Ma$^2$\quad\quad Madian Khabsa$^2$ \\
$^1$Hong Kong University of Science and Technology \quad\quad $^2$Facebook AI \\
\texttt {nayeon.lee@connect.ust.hk} \\
\texttt {\{belindali,sinongwang,scottyih,haom,mkhabsa\}@fb.com}
}

\date{}

\begin{document}
\maketitle
\begin{abstract}
Recent work has suggested that language models (LMs) store both common-sense and factual knowledge learned from pre-training data. In this paper, we leverage this implicit knowledge to create an effective end-to-end fact checker using a \textit{solely} a language model, without any external knowledge or explicit retrieval components.
While previous work on extracting knowledge from LMs have focused on the task of open-domain question answering, to the best of our knowledge, this is the first work to examine the use of language models as \textit{fact checkers}.
In a closed-book setting, we show that our zero-shot LM approach outperforms a random baseline on the standard FEVER task, and 
that our finetuned LM compares favorably with standard baselines.
Though we do not ultimately outperform methods which use explicit knowledge bases, we believe our exploration shows that this method is viable and has much room for exploration.
\end{abstract}


\section{Introduction}
Pre-trained language models have recently 
lead to significant advancements in wide variety of NLP tasks, including question-answering, commonsense reasoning, and semantic relatedness \cite{devlin2018bert,radford2019language,peters2018deep,radford2018improving}. 
These models are typically trained on documents mined from Wikipedia (among other websites). 
Recently, a number of works have found that LMs store a surprising amount of world knowledge, focusing particularly on 
the task of open-domain question answering  \cite{petroni2019language,roberts2020much}. In this paper, we explore
whether we can leverage the knowledge in LMs for \textit{fact checking}.

\begin{figure}[t]
	\centering
	\begin{subfigure}[t]{\linewidth}
		\centering
		\includegraphics[width=\linewidth]{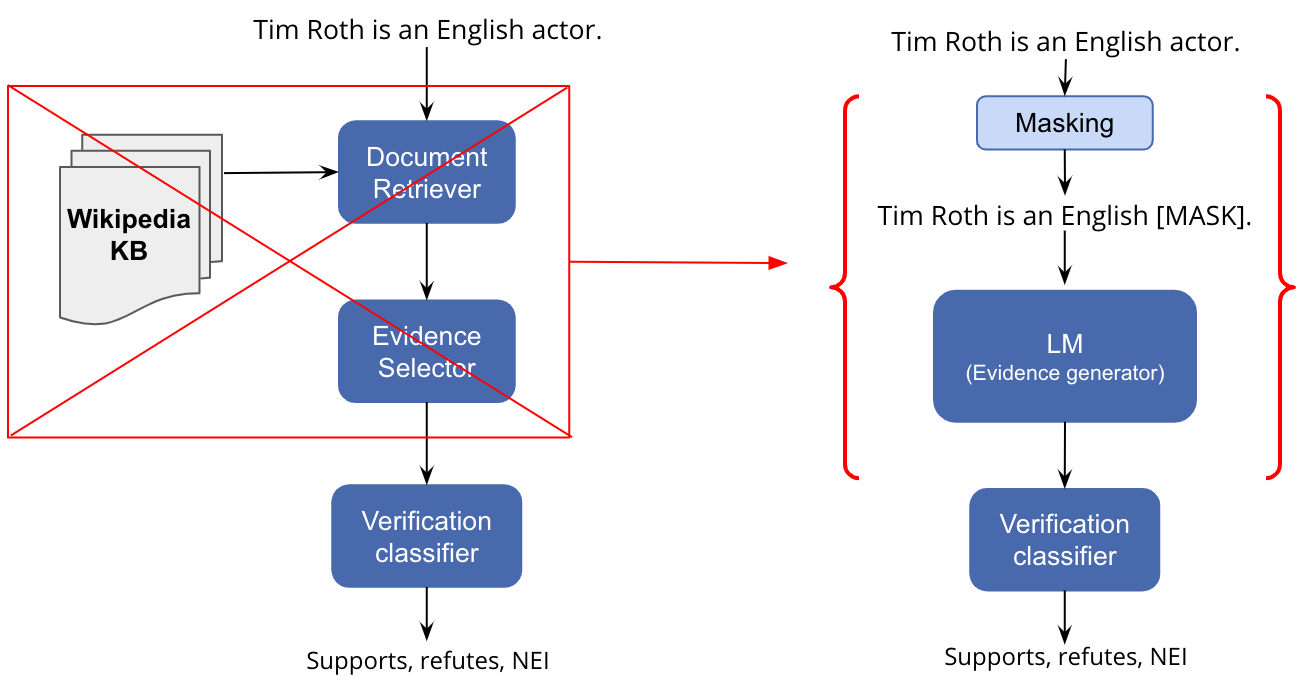}
	\end{subfigure}
	\begin{subfigure}[t]{1.5in}
		\centering
		\caption{Traditional fact-checking pipeline.}\label{fig:diagram_a}	
	\end{subfigure}
	\quad
	\begin{subfigure}[t]{1in}
		\centering
		\caption{Our new fact-checking pipeline.}\label{fig:diagram_b}
	\end{subfigure}
	\caption{Traditional fact-checking pipeline (left) vs. Our LM-based pipeline (right)}\label{fig:diagram}
\end{figure}

We propose an approach (Fig. \ref{fig:diagram_b}) that replaces the document retriever and evidence selector models in traditional fact-checking (Fig. \ref{fig:diagram_a}) with a single language model that generates 
masked tokens. 
This offers a number of advantages over the traditional approach: first, the procedure is overall simpler, requiring fewer resources and computation -- we do not need to maintain an explicit knowledge base external to our LM, and we do not need an explicit retrieval step. The latter in particular can lead to a 
huge speedup in the system, since we can skip the time-consuming step of searching over a potentially massive space of documents. Second, LMs are widely-available and are currently attracting significant research effort. 
Thus, research in language-modeling, particularly in improving LMs ability to memorizing knowledge, may also improve the overall effectiveness of our fact-checking pipeline. 
Lastly, our system further shifts the paradigm towards ``one model for all" --- LMs have been used for a wide variety of tasks, and now also for fact checking. 


In order to determine the feasibility of our approach, we start
with a human review study where participants are given a claim
from FEVER \cite{thorne2018fever}, and are asked to validate the claim using only a BERT language model. 
We found that users had reasonable success in determining 
claim validity. Empowered by the results, we design an \textit{end-to-end} neural approach for utilizing BERT as a fact checker (see Figure \ref{fig:diagram_b}). 
At a high level, we first generate an evidence sentence by masking the claim and using BERT to ``fill in" the mask. We then feed the generated sentence, alongside the original claim, to a verification classifier model that classifies whether the claim is supported, refuted, or the information is insufficient to make a call.

The rest of the paper is organized as such: Section \ref{sec:background} gives an overview of the problem space. Section \ref{sec:human_experiment} describes our preliminary experiments. Sections \ref{sec:model} and \ref{sec:experiment} highlights our main methods (i.e. end-to-end model, experimental setup), and \ref{sec:results} reports our main results. Sections \ref{sec:analysis} and \ref{sec:conclusion} conclude our paper with a discussion and future works.

\section{Background}
\label{sec:background}
\paragraph{Task} The main goal of fact-checking is to validate the truthfulness of a given claim. Each claim is assigned one of three labels: support, refute, or not enough information (NEI) to verify.

\paragraph{Dataset} We use FEVER~\cite{thorne2018fever}, a large-scale fact-checking dataset with around 5.4M Wikipedia documents. Claims were generated by extracting sentences from Wikipedia (with possible mutations), and were annotated by humans with 
their verification label and/or evidence sentences from Wikipedia.

\paragraph{Traditional pipeline} Traditional fact-checking pipelines \cite{malon2018,hanselowski2018ukp,yoneda2018ucl,lee2018improving,nie2019combining} access knowledge within an external knowledge base (i.e. Wikipedia) to validate a claim (Fig. \ref{fig:diagram_a}). They use a multi-step, pipe-lined approach, which involve IR-modules, such as document retrievers and evidence selectors, for retrieving the appropriate evidence, and verification modules that take in $\texttt{\{claim, [evidences]\}}$ pairs and predict a final verification label.

\paragraph{Proposed pipeline} As shown in Fig.\ref{fig:diagram_b}, our proposed pipeline replaces both the external knowledge base as well as the IR modules with a pretrained language model. In the remainder of this paper, we utilize BERT. Future work can explore other language models.

\paragraph{Querying the Language Model}
In \newcite{petroni2019language}, language models
were used as knowledge base to answer open-domain questions.
To do this, the authors devised a probe known as ``LAMA", which generates fill-in-the-blank cloze-style statements from questions. For example, in order to answer the question `Where is Microsoft's headquarter?', the question would be rewritten as as `Microsoft's headquarter is in [MASK]' and fed into a language model for the answer. Inspired by LAMA \cite{petroni2019language}, we generate evidences from language models through fill-in-the-blank style tasks. 

\newcite{jobanputra2019unsupervised} attempted to leverage the fill-in-the-blank style for fact-checking on a simplified FEVER setup (i.e. “supports” vs “manual review”). All the named-entities were masked, and the ratio of correctly reconstructed named entities over all masked named entities was used for label classification. Our approach, on the other hand, tries to consciously choose which token to mask, introduces an entailment model to capture the semantic relation between the claim and the evidence, and conducts experiment in the original FEVER setup. 

\section{Exploratory Experiments}
\label{sec:human_experiment}
In order to determine the feasibility of our approach, we began by conducting
a human review study on 50 random-selected claims from FEVER \cite{thorne2018fever}. Participants were asked to validate each claim with \textit{only} a language model, by following these steps:  

\begin{enumerate}
    \item Mask a token from the claim, depending on component of the claim we wish to verify: \\
    \textbf{ 
    Thomas Jefferson founded the University of \underline{Virginia} after retiring} $\to$  \textbf{Thomas Jefferson founded the University of \textbf{[MASK]} after retiring}. \\
    In this example, the user is verifying which university
    was founded by Thomas Jefferson. Note that the user could alternatively choose to mask  \textit{Thomas Jefferson} in order to verify the founder of University of Virginia.
    \item Get the top-1 predicted token from the LM. \\ \textbf{Top-1 predicted token = Virginia}. 
    \item If predicted token matches the masked token, the claim is supported, otherwise it is refuted. \\
    \textbf{Virginia $\equiv$ Virginia $\to$ SUPPORTS}
\end{enumerate}

In other words, we asked participants to serve as the ``masking" and ``verification classifier" components of our fact-checking pipeline in Fig. \ref{fig:diagram_b}.

Two participants examined the 50 claims, and eventually achieved an average accuracy of $55\%$. \footnote{Both participants had NLP background, and both were familiar with FEVER and the fact-checking task. We also assumed both participants were capable of selecting the optimal position to mask.}


We also conducted this zero-shot study on a larger scale and in a more systematic way, by taking all claims in the \textit{full} FEVER dataset, and always masking the last token.\footnote{We omit examples for which the masked token is not in BERT's vocab.}
Otherwise, we preserve steps 2 and 3 from above.
Even with this na\"{i}ve token-matching approach, we were able to obtain precision $56\%$ and F1 $59\%$ for the positive label (SUPPORT).

Our preliminary experiments' results illustrate that, with a good masking mechanism and verification model, language models can indeed feasibly be used for fact-checking.

\section{End-to-End Fact-Checking Model}
\label{sec:model}

\begin{table*}[t]
\small
\resizebox{\textwidth}{!}{%
    \begin{tabular}{lcccccccc}
    \toprule
    \textbf{Model} & \textbf{Label} & \textbf{prec} & \textbf{recall} & \textbf{f1} & \textbf{accuracy} & \textbf{macro prec} & \textbf{macro recall} & \textbf{macro f1} \\ \midrule
    \multirow{3}{*}{$BERT_{freeze}$} & REFUTES & 0.36 & 0.69 & 0.47 & \multirow{3}{*}{0.38} & \multirow{3}{*}{0.39} & \multirow{3}{*}{0.38} & \multirow{3}{*}{0.33} \\
     & SUPPORTS & 0.43 & 0.09 & 0.15 &  &  &  &  \\
     & NEI & 0.39 & 0.35 & 0.37 &  &  &  &  \\ \midrule
    \multirow{3}{*}{$BERT_{finetune}$} & REFUTES & 0.62 & 0.55 & 0.58 & \multirow{3}{*}{0.57} & \multirow{3}{*}{0.57} & \multirow{3}{*}{0.57} & \multirow{3}{*}{0.57} \\
     & SUPPORTS & 0.54 & 0.67 & 0.59 &  &  &  &  \\
     & NEI & 0.57 & 0.49 & 0.53 &  &  &  &  \\ \midrule
     \multirow{3}{*}{$BERTasKB$} & REFUTES & 0.76 & 0.38 & 0.51 & \multirow{3}{*}{0.49} & \multirow{3}{*}{0.59} & \multirow{3}{*}{0.49} & \multirow{3}{*}{0.44} \\
     & SUPPORTS & 0.41 & 0.92 & 0.57 &  &  &  &  \\
     & NEI & 0.58 & 0.15 & 0.24 &  &  &  &  \\ \midrule
     Shared-Task Baseline \cite{thorne2018fact} * & - & - & - & - & \textbf{0.49} & - & - & - \\ 
     Shared-Task SoTA \cite{thorne2018fact} * & - & - & - & - & \textbf{0.68} & - & - & - \\ 
     DREAM SoTA \cite{zhong2019reasoning} * & - & - & - & - & \textbf{0.77} & - & - & - \\ \bottomrule
    \end{tabular}%
}
\label{table:results}
\caption{Performance comparison between BERT-as-encoder models
($BERT_{freeze}$, $BERT_{finetune}$) and BERT-as-LM model ($BERTasKB$). New SoTA result (DREAM) released in ACL2020 is added for reference (*We report fact-checking label accuracy, not the stricter FEVER score) }
\end{table*}

Enlightened by results from our preliminary experiments, we devise an end-to-end model that
automates and improve upon the masking and verification steps that were conducted by humans.
Specifically, we resolve two limitations: 1. manual masking of claims, and 2. na\"{i}ve validation of the predicted token that fails to deal with synonyms and other semantic variants of the answer.

\paragraph{Automatic Masking} 
We mask the \textit{last named entity} in the claim, which we identify using an off-the-shelf Named-Entity-Recognition (NER) model from spaCy \newcite{honnibal2017spacy}. In particular, we choose to mask named entities in order to better ensure that the token we mask actually makes use of the \textit{knowledge} encoded in language models. (Otherwise, we may mask tokens that only make use of the LM's ability to recover linguistic structures and syntax -- for instance, masking stopwords). This hinges on the observation that, for most claims, its factuality hinges upon the correctness of its entities (and the possible relations between them), and \textit{not} on how specifically the claim is phrased.


\paragraph{Verification using Entailment} To move beyond na\"{i}vely matching predicted and gold tokens, we leverage a textual entailment model from AllenNLP \cite{gardner2018allennlp} to validate our LM predictions. Note that textual entailment models predict the directional truth relation between a text pair (i.e. ``sentence \textit{t} entails \textit{h}" if, typically, a human reading \textit{t} would infer that \textit{h} is most likely true).

\begin{figure}[t]
    \small
    \centering
    \includegraphics[width=\linewidth]{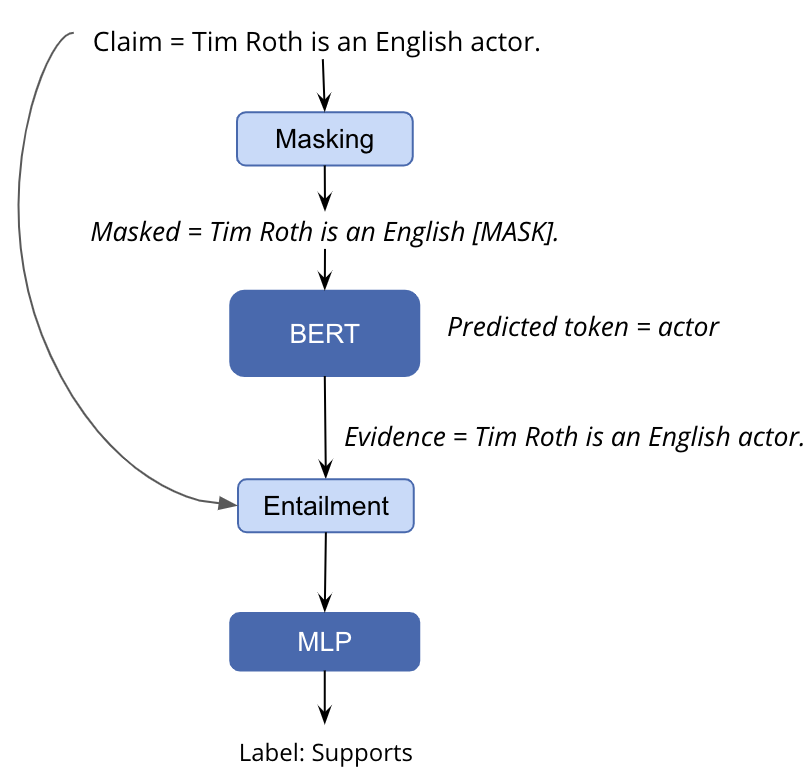}
    \caption{Detailed illustration of our pipeline}
    \label{figure:new_pipeline}
\end{figure}

\paragraph{Full-pipeline steps} Detailed steps for our end-to-end model (Fig. \ref{figure:new_pipeline}) are as follows:
\begin{enumerate}
    \item Masked the last named entity found by the NER model. 
    \item Get the top-1 predicted token from the LM, and fill in the [MASK] accordingly to create the ``evidence'' sentence. 
    \item Using the claim and generated ``evidence'' sentence, obtain entailment ``features" using outputs from the last layer of the pretrained entailment model (before the softmax).
    \item Input the entailment features into a multi-layer perceptron (MLP) for final fact-verification prediction.
\end{enumerate}

\section{Experiments}
\label{sec:experiment}
\subsection{Experiment setup}
We conduct our experiments on the FEVER claim verification
dataset \cite{thorne2018fever} using the standard provided splits.
We use the publicly available 24-layer BERT-Large as our language model, which was pre-trained on Wikipedia in 2018.\footnote{It's possible the model was trained on a later Wikipedia dump than what's released as part of FEVER, but pre-training BERT from scratch is beyond the scope of this paper.}

The MLP was optimized using Adam, and trained with a mini-batch size of $32$. The learning rate was set to $0.001$ with max epoch size $200$ and epoch patience of $30$. The embedding size of the entailment features (from the pre-trained entailment model) was $400$, and our MLP classifier had hidden size of $100$.

\subsection{Evaluation Metric}
The traditional pipeline was evaluated using FEVER scoring, which is a stricter form of scoring that treats predictions to be correct only when correct evidences were retrieved. Since our pipeline does not utilize an external knowledge base, and does not have an evidence retriever, we only examine the correctness of the final verification step using precision, recall, F1 and accuracy. We leave generating evidences with language models 
for future work.

\subsection{Baselines}
We introduce two language model baselines for comparison. 
The first baseline, $BERT_{freeze}$, uses an MLP layer on top of a frozen BERT encoder 
to make predictions (gradients backpropagate to the MLP layer only).
In this baseline, we aim to \textit{extract} the already 
stored knowledge within BERT model as an embedding vector, and avoid finetuning 
the internal layers, in order to disentangle BERT's knowledge 
from it's ability to serve as a high-capacity classifier.

The second baseline, $BERT_{finetune}$, allows all the model layers to be updated 
based on the fact-verification loss from the MLP layer. This baseline captures
BERT's ability as \textit{both} a language model, and a high-capacity text encoder. 

\begin{table*}[t]
\small
\centering
\begin{tabular}{clccc}
\toprule
\textbf{ID} & \textbf{Claim} & \textbf{Masked Token} & \textbf{Predicted Token} & \textbf{Label} \\ \midrule
$a$ & Kuching is the capital of {[}MASK{]}. & Sarawak & Sarawak & SUPPORTS \\
$b$ & The Beach's director was Danny {[}MASK{]}. & Boyle & Boyle & SUPPORTS \\ \midrule
$c$ & Tim Roth was born in {[}MASK{]} & 1961 & London & SUPPORTS \\
$d$ & Chile is a {[}MASK{]}. & country & democracy & SUPPORTS \\
$e$ & Seohyun {[}MASK{]}. & sings & Park & SUPPORTS \\ \bottomrule
\end{tabular}
\caption{Examples of token predictions from BERT in zeroshot setting. $a$, $b$ are correctly fact-checked examples, and $c$, $d$, $f$ are wrongly fact-checked examples.}
\label{table:token_predictions}
\end{table*}

Note that the dataset is evenly distributed
among the three classes, therefore a random baseline
would yield an accuracy of 33\%. Also note that the Fever-baseline model introduced by the task organizers achieves accuracy score of 48.8\% \cite{thorne2018fact}.

\section{Results and Discussion}
\label{sec:results}
The results of the three models are reported in Table 1. We 
observe that our proposed approach ($BERTasKB$) outperforms 
$BERT_{freeze}$ on all metrics suggesting that querying 
language models in QA style is a better approach 
for extracting their encoded knowledge. 
Similarly, $BERTasKB$ model achieves an accuracy score of 
49\% which is comparable to Fever-baseline at 48.8\%, except without
the need for explicit document retrieval and evidence selection.
This suggests that language models, used as sources of knowledge
for fact checking, are at least as effective as standard baselines.
However, there is still much room for future research, as the state-of-the-art model on the Fever shared task achieves an accuracy score of 68.21\% \cite{thorne2018fact}. 

On the other hand, we find that $BERTasKB$ lags behind $BERT_{finetune}$, as expected,
on most metrics. We hypothesize this is due to the high capacity of the model, in comparison,
and to the effectiveness of BERT models in text classification. 
Upon examining the results of these two models closely, we find that 
$BERTasKB$ struggles mightily with the \texttt{NEI} category (F1 score of
0.24 vs 0.53) indicating that our current approach might need 
specific modules to better tackle that category. As both models 
seem to be equally adept in identifying the  \texttt{support} class 
(0.57 vs 0.59 F1), indicating that $BERTasKB$ is unable to 
distinguish between \texttt{refute} and \texttt{NEI} classes.
Future work can further investigate techniques to identify these two 
categories. 
Interestingly, the 
$BERT_{freeze}$ achieves an accuracy score of 38\% which is 
slightly better than a random baseline which achieves 33\%.

\section{Analysis of Token Prediction Results}
\label{sec:analysis}
In this section, we provide some examples of tokens predicted from BERT to understand the performance of ``evidence generation''. 

First two examples in Table \ref{table:token_predictions} ($a$, $b$) are examples with correct fact-check labels from zeroshot setting. When a claim has enough context, and contains rather rare names such as ``Sarawak'', BERT manages to predict correct tokens. 

We also provide detailed analysis on the error cases to facilitate future work in making further improvements: 
\begin{itemize}
    \item One common form of errors is that, the entity type of token prediction is biased towards the way how the training data was written. For example, sentence $c$ from Table~\ref{table:token_predictions} illustrates a common claim structure in FEVER dataset which talks about the birth-year of a person (e.g., Tim Roth). However, $100\%$ of our test samples with such structure always predict city/country (e.g., London). The reason is, in Wikipedia, the birth-years are always written in the following structure ``PERSON (born DATE)'' (e.g., ``Tim Roth (born 14 May 1961)''), and birth city/country written in ``PERSON was born in city/country'' structure (e.g., ``Roth was born in Dulwich, London''). Therefore, to obtain birth-year, the claim had to be written as Tim Roth (born [MASK]) to predict correctly.
    \item Sentence $d$ is another example that the entity type of token prediction is hard to control. ``is a...'' is a very general prefix phrase, making it hard for BERT model to correctly predict correct entity type. 
    \item There are lots of short claims in FEVER test set (approx. $1100$ samples) which has less than 5 tokens (e.g. sentence $e$). Since there is very little context, BERT struggles to predict correctly. 
\end{itemize}

One of the the main insight we get from these analysis is that, the way the language model is initially pre-trained, greatly determines the way it should be ``queried''.

\section{Conclusions \& Future Work}
\label{sec:conclusion}
In this paper, we explored a new fact-checking pipeline that use language models as knowledge bases. Unlike previous pipelines that required dedicated components for document retrieval and sentence scoring, our approach simply translates a given claim into a fill-in-the-blank type query and relies on a BERT language model to generate the ``evidence''. Our experiment shows that this approach is comparable to the standard baselines on the FEVER dataset, though not enough to beat the state-of-the-art using the traditional pipeline. However, we believe our approach has strong potential for improvement, and future work can explore using stronger models for generating evidences, or improving the way how we mask claims. 

In the future, we will investigate sequence-to-sequence language models such as BART \cite{lewis2019bart} or T5 \cite{raffel2019exploring}, that have recently shown to be effective on generative question-answering \cite{roberts2020much}. Similarly, our proposed approach seem to struggle with correctly identifying NEI cases, and we plan to investigate adding specific modules to deal with NEI. Lastly, we plan to explore new ways of pre-training language models to better store and encode knowledge.

\section*{Acknowledgements}
We would like to thank Fabio Petroni for the helpful discussion and inspiration.

\bibliography{acl2020}
\bibliographystyle{acl_natbib}

\end{document}